\documentclass[a4paper, 10pt, conference]{ieeeconf}  

\IEEEoverridecommandlockouts            

\usepackage{graphicx} 
\usepackage{hyperref}
\hypersetup{hidelinks}
\usepackage{svg}
\usepackage{amsmath} 
\usepackage{amssymb}  
\usepackage{float}

\title{\LARGE \bf
Design and Evaluation of a Bioinspired Tendon-Driven 3D-Printed Robotic Eye with Active Vision Capabilities}

\author{Hamid Osooli$^{1}$, Mohsen Irani Rahaghi$^{2}$ and S. Reza Ahmadzadeh$^{1}$
\thanks{$^{1}$Persistent Autonomy and Robot Learning (PeARL) Lab, University of Massachusetts Lowell, Lowell, MA 01854, USA
        {\tt\small \{hamid\_osooli, reza\_ahmadzadeh\}@uml.edu}}%
\thanks{$^{2}$Mechanical Engineering Department, University of Kashan, Kashan, Isfahan, Iran {\tt\small irani@kashanu.ac.ir}}}

\begin{document}
\maketitle
\thispagestyle{empty}
\pagestyle{empty}

\begin{abstract}
The field of robotics has seen significant advancements in recent years, particularly in the development of humanoid robots. One area of research that has yet to be fully explored is the design of robotic eyes. In this paper, we propose a computer-aided 3D design scheme for a robotic eye that incorporates realistic appearance, natural movements, and efficient actuation. The proposed design utilizes a tendon-driven actuation mechanism, which offers a broad range of motion capabilities. The use of the minimum number of servos for actuation, one for each agonist-antagonist pair of muscles, makes the proposed design highly efficient. Compared to existing ones in the same class, our designed robotic eye comprises aesthetic and realistic features. We evaluate the robot's performance using a vision-based controller, which demonstrates the effectiveness of the proposed design in achieving natural movement, and efficient actuation. The experiment code, toolbox, and printable 3D sketches of our design have been open-sourced.
\end{abstract}

\section{INTRODUCTION}

Despite the advancements in robotic technology, replicating the capabilities and movements of the human eye has remained a topic of active research in recent years. To better understand the mechanics and functionalities of the human eye, several studies focus on creating robots that mimic its movements and abilities~\cite{rajendran2021two, osooli2019game, rajendran2022observability}. 
  
Previous efforts to design a mechanism similar to the human eye have been based on mechanical models, such as the Ruete mechanical model, also known as the Ophtalmotrope~\cite{b8pruehsner2006davinci, b7simonsz199019th}.
The Ophtalmotrope in~\cite{b8pruehsner2006davinci} features two wooden eyeballs, each linked to six polypropylene Masons Strings which represent the extra-ocular muscles. 

In recent decades, researchers have designed robotic eyes by utilizing advanced materials such as Super Coiled Polymers~\cite{rajendran2021two}, Folded Dielectric Elastomer Actuators~\cite{b16carpi2007bioinspired} and Pneumatic Artificial Muscles~\cite{b17wang2008design}. The disadvantage of using advanced materials is that the system requires additional components to measure the material properties which increases the complexity of the whole system. For instance measuring the electric current passing through the super coiled polymer muscles requires an extra Raspberry Pi~\cite{rajendran2021two}. Another example is measuring the orientation for the Pneumatic Artificial Muscles by an embedded 3-Axis Micro Electro Mechanical accelerometer in~\cite{b17wang2008design}. 
On the other hand, robotic eyes that are too simplified, lack aesthetics and appearance~\cite{b11wang2013advanced, b15schultz2012camera}. 

Efficient actuation and natural movements are among the main features of an effective human-like robotic eye. One technique towards creating human-like movements in robots is the use of tendon-driven mechanisms for robot actuation. Tendon-driven robots, in which wires (tendons) serve as muscle-like elements, provide greater freedom of movement compared to traditional, rigid systems~\cite{b2mizuuchi2002design}. The use of tendon-driven systems for various robotic applications has been explored by researchers~\cite{b3kobayashi1998tendon, b5takuma2018body, lin2022modular}. However, there has been limited research on the application of tendon-driven systems to the design of the robots that mimic the mechanism of the human eye. The drawback of the existing works that use tendon-driven mechanism is the use of redundant actuators in the design~\cite{rajendran2021two, b6biamino2005mac}.

In this paper we propose a bioinspired design for a robotic eye that uses a minimum number of actuators to achieve natural human-like movements. Our designed system replicates human eye movements, while also using a minimum number of actuators. We equip our robot with a vision-based controller and evaluate its performance over the main four movements of human eye: Saccadic, Smooth Pursuit, Vergence, and Vestibulo-Ocular Reflex (VOR).

We perform eight experiments (two per movement) to showcase the capabilities of the proposed robot. Our results show that the proposed robot is capable of mimicking all four categories of human eye movements by detecting and following human face while centralizing it simultaneously. The code for the experiments, toolbox, and the printable 3D sketches of the proposed robotic eye are open sourced.

\section{Background}
\subsection{The Human Eye}

The human eye is a roughly spherical organ that is composed of two main parts: the cornea and the sclera. The cornea is the front part of the eye that is more curved, and includes the black center of the eye (the pupil) and the colored ring around it (the iris). The sclera is the white part of the eyeball that contains the necessary structures for receiving visual information. In order to see clear images, the image must be focused on a small region called the fovea located within the sclera. As a result, the eye must move to bring an image into focus on the fovea~\cite{b19perkins2015human}.

Six extra-ocular muscles are responsible for rotating each eye within the bony orbit of the skull. These muscles are composed of four rectus muscles (superior, inferior, medial, and lateral) and two oblique muscles (superior and inferior). The rectus muscles attach to the eyeball at a point 55 degrees away from the optical axis~\cite{b20miller1984model}, and are responsible for moving the eye in the horizontal and vertical directions. The oblique muscles pass through a structure called the trochlea, and rotate the eye around the visual axis. To move the eye, a pair of these muscles work in an agonist-antagonist fashion; for example, the medial rectus muscle contracts while the lateral rectus muscle relaxes to rotate the eye towards the nose.

\subsection{Eye Movements}

There are several motivations for eye movement. One of the main motivations is to fixate on a target in the visual field. Another motivation is to prevent visual adaptation to a static scene, which can lead to a phenomenon known as neural fading~\cite{b22krekelberg2011microsaccades}. This can be prevented by small, involuntary eye movements called micro Saccades that refresh the visual input~\cite{b23KOWLER1980273}. Eye movements also continue to occur during sleep, specifically during Rapid Eye Movement (REM) sleep~\cite{b24horne2013rem}.

There are four main categories of eye movements: Saccades, Smooth Pursuit, Vergence, and Vestibulo-Ocular Reflex (VOR). Saccades are quick, jerky movements of the eyes that bring the gaze to a new target. For example, tracking a flying insect in the air. Smooth pursuit movements are controlled movements of the eyes that are used to track a moving target smoothly, such as when reading a line of text. Vergence movements involve the simultaneous movement of the eyes in opposite directions towards each other, which occurs when an object is located at close proximity. VOR, is an oculomotor reflex that compensates for head movements by moving the eyes in the opposite direction of the head movement. It is dependent on the vestibular system, which is responsible for detecting head movement. 


\section{Bioinspired Design}

In this paper, we focus on the development of a robotic eye by designing and building an accurate mechanical model of the human eye. The resulting robotic eye could be used in several robotics and vision applications such as stereo cameras, humanoid robots, and active vision. To achieve this goal, we leverage Computer Aided Design (CAD) to construct a robotic eye. The benefit of using CAD is that we are able to 3D print the mechanical sketches using Acrylonitrile Butadiene Styrene (ABS) filaments. In this section, we provide a detailed description of the design considerations for each component of our proposed robotic eye, including the cornea, iris, and lens. We also explain how we validate the accuracy of the model and how it can be used in future research studies. 

\subsection{The Eyeball}

\begin{figure}[t]
    \centering
    \includegraphics[width=\columnwidth]{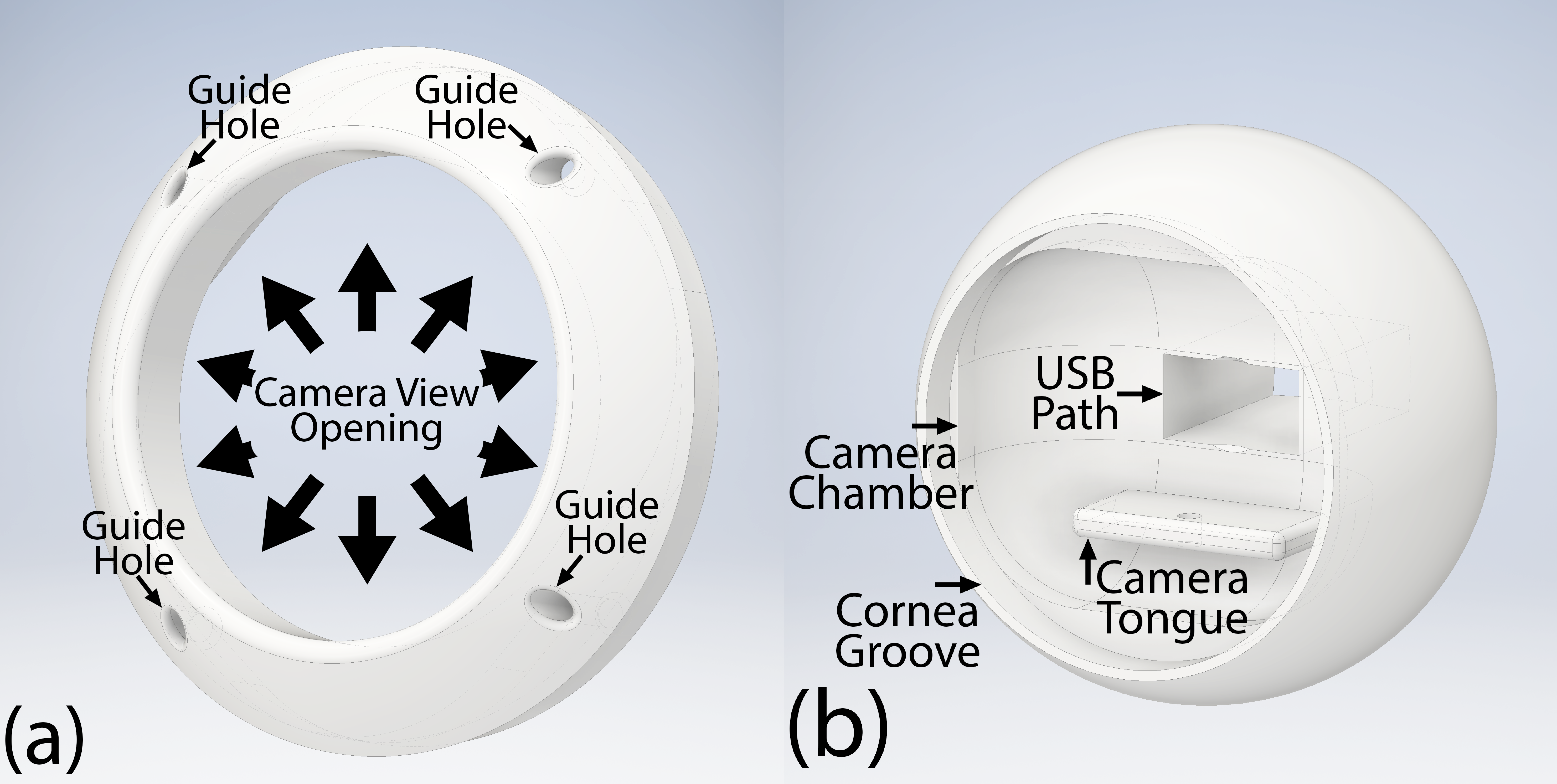}
    \caption{The designed cornea (a) and sclera (b). a illustrates the camera view opening and guide holes, while b features the camera chamber, integrated groove, and cable output}
    \label{fig:eyeball}
\end{figure}

The mechanical eyeball consists of a chamber, shown in Fig.~\ref{fig:eyeball}, to host a  single-view USB camera. We selected a Microsoft LifeCam HD-3000 webcam as it is an inexpensive and readily available perception sensor that helps keep the overall cost of the robot low.

To achieve this goal, we designed a 60 mm diameter eyeball that could accommodate our USB camera. This is more than twice the size of a typical adult human eyeball, which varies from 21 mm to 27 mm~\cite{b25bekerman2014variations}. First, we designed the sclera, which serves as the outermost layer of the robotic eye, and houses the camera. As shown in Fig.~\ref{fig:eyeball} the sclera has an integrated groove to hold the camera and a USB path to pass the USB cable.

Next, we add the cornea to complement the body of the eyeball. The cornea connects to the sclera at a 55 degrees angle with respect to the light axis, which allows the tendon-driven cords to move the eyeball in a manner similar to the human eye~\cite{b20miller1984model}. The cornea has an opening with the maximum radius possible without obstructing the camera view, and four guide holes on the cornea help to tighten the cords over the eyeball. Details of the parts are shown in Fig.~\ref{fig:eyeball}.

\subsection{The Orbit}

\begin{figure}[t]
    \centering
    \includegraphics[width=\columnwidth]{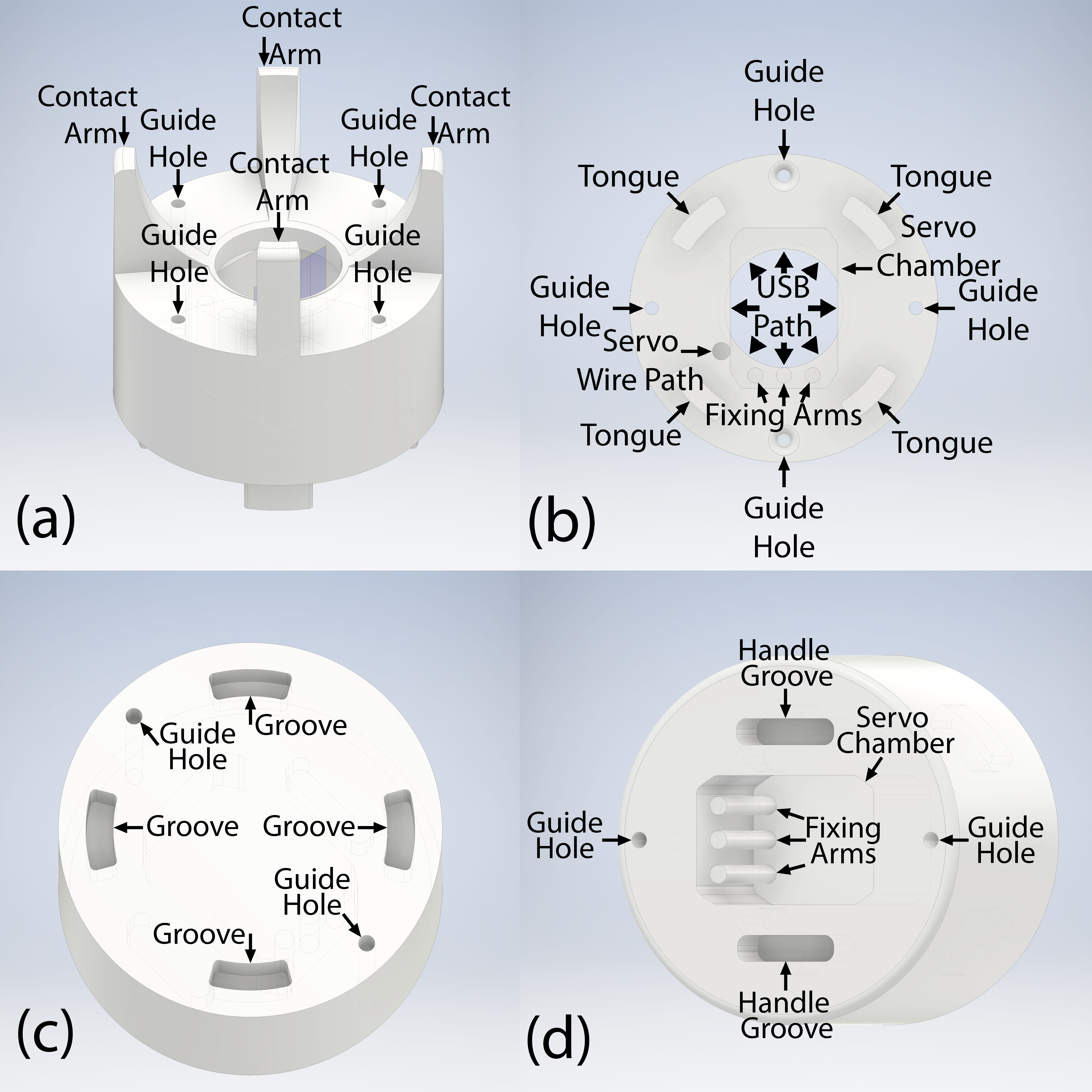}
    \caption{The proposed robotic eye's orbit sections including the front (a)-(b), and back (c)-(d) components}
    \label{fig:orbit}
\end{figure}

Another key component of our robotic eye is the orbit. In human eye it is the skull cavity in which the eye and extra-ocular muscles are placed. Since we are only designing a mechanical model of the eye and not a complete head, we propose a bi-sectional orbit design to separate the chambers that host the servos. The details of the parts can be seen in Fig.~\ref{fig:orbit}.

As shown in Fig.~\ref{fig:orbit}(b) and~\ref{fig:orbit}(d), each section of the orbit hosts a servo that actuates the robotic eye's movement through cords similar to the Rectus muscles in the human eye. The cords are responsible for movement in the horizontal and vertical directions. We designed one actuator for each agonist-antagonist pair of extra-ocular muscles.

The front section of the orbit (Fig.~\ref{fig:orbit}(a) and Fig.~\ref{fig:orbit}(b)) has four contact arms placed in every 75 degrees. The design method and polishing of the eyeball surface decreases friction between the eyeball and orbit. The front section also includes a path for the USB cable of the camera, and four guide holes that cross the cords. The servo chamber behind this section hosts the vertical actuator and stabilizes it with three fixing arms. The front section connects to the orbit back section with a tongue and groove joint.

The back section of the orbit as shown in Fig.~\ref{fig:orbit}(c) and Fig.~\ref{fig:orbit}(d), hosts one horizontal actuator, and connects to the handle (details in the next section) with tongue and groove joint. This allows the user to hold and control the robot with ease.

\subsection{The Handle}

\begin{figure}[t]
\centerline{\includegraphics[trim=7cm 7cm 32cm 22cm, clip, width=\columnwidth]{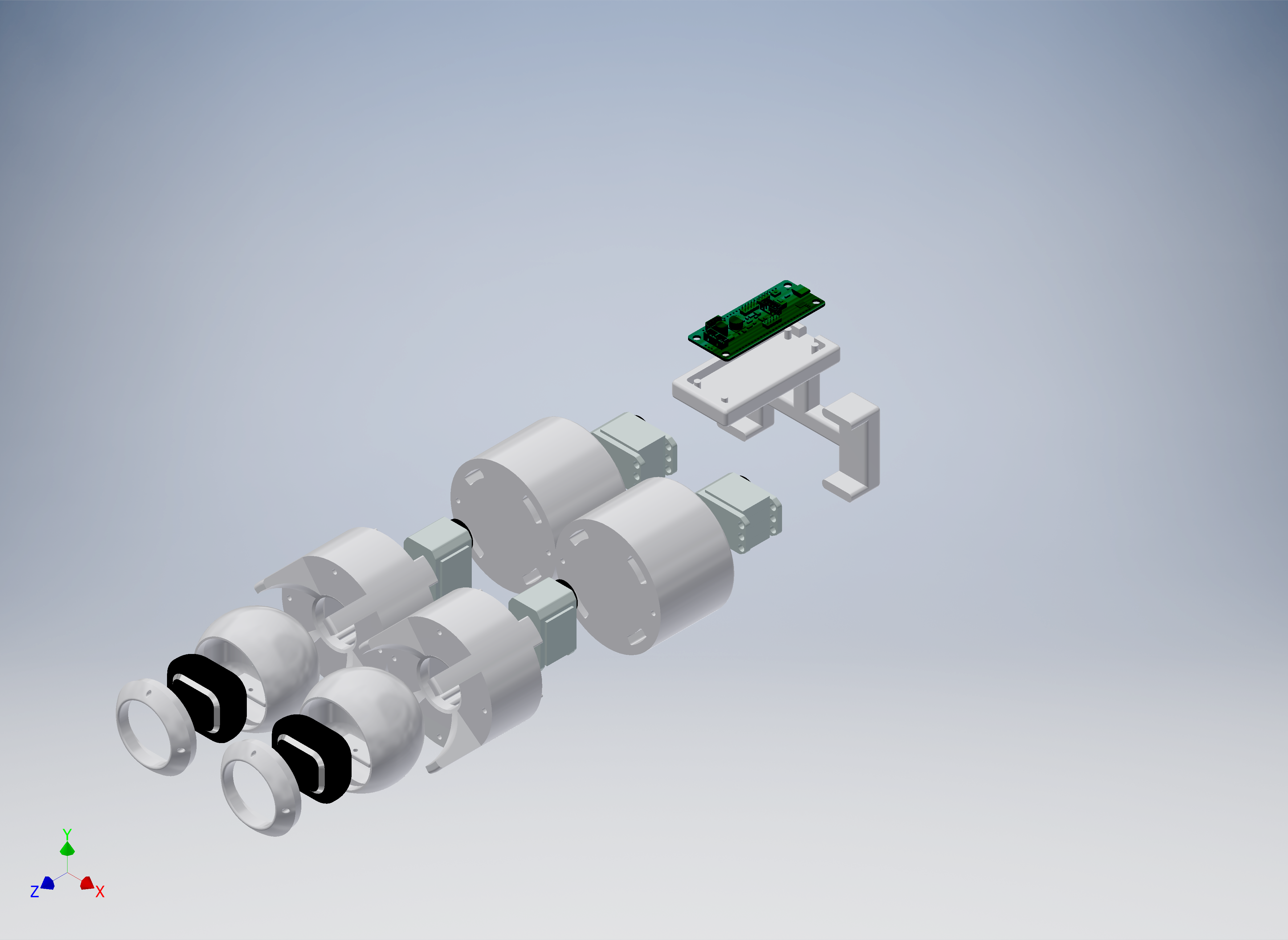}}
	\caption{Exploded view of the proposed robotic eye including the 3D-printed body, two Microsoft HD-3000 cameras, four Dynamixel XL-320 servos, and the OpenCM 9.04-C controller board.}
	\label{fig:exploded}
\end{figure}

As shown in Fig.~\ref{fig:exploded}, the final component of the robotic eye is the handle, which provides multiple functionalities. Its primary functionality is to keep the eyes' locations fixed in a 7 cm distance from each other, which is similar to the human eyes' distance, that is about 6.3 cm~\cite{b26dodgson2004variation}. Considering this distance helps to create a more realistic mechanical model of the human eye, and is practical when using the robot for stereo vision.

The handle is designed to be compact and easy to hold, it includes a chamber for the controller board which interacts with the PC and the servos. The handle connects to the robot using a tongue and groove joint, which ensures a secure and stable connection.

\begin{figure}[t]
\centerline{\includegraphics[trim=0cm 10cm 0cm 0, clip, width=.5\columnwidth]{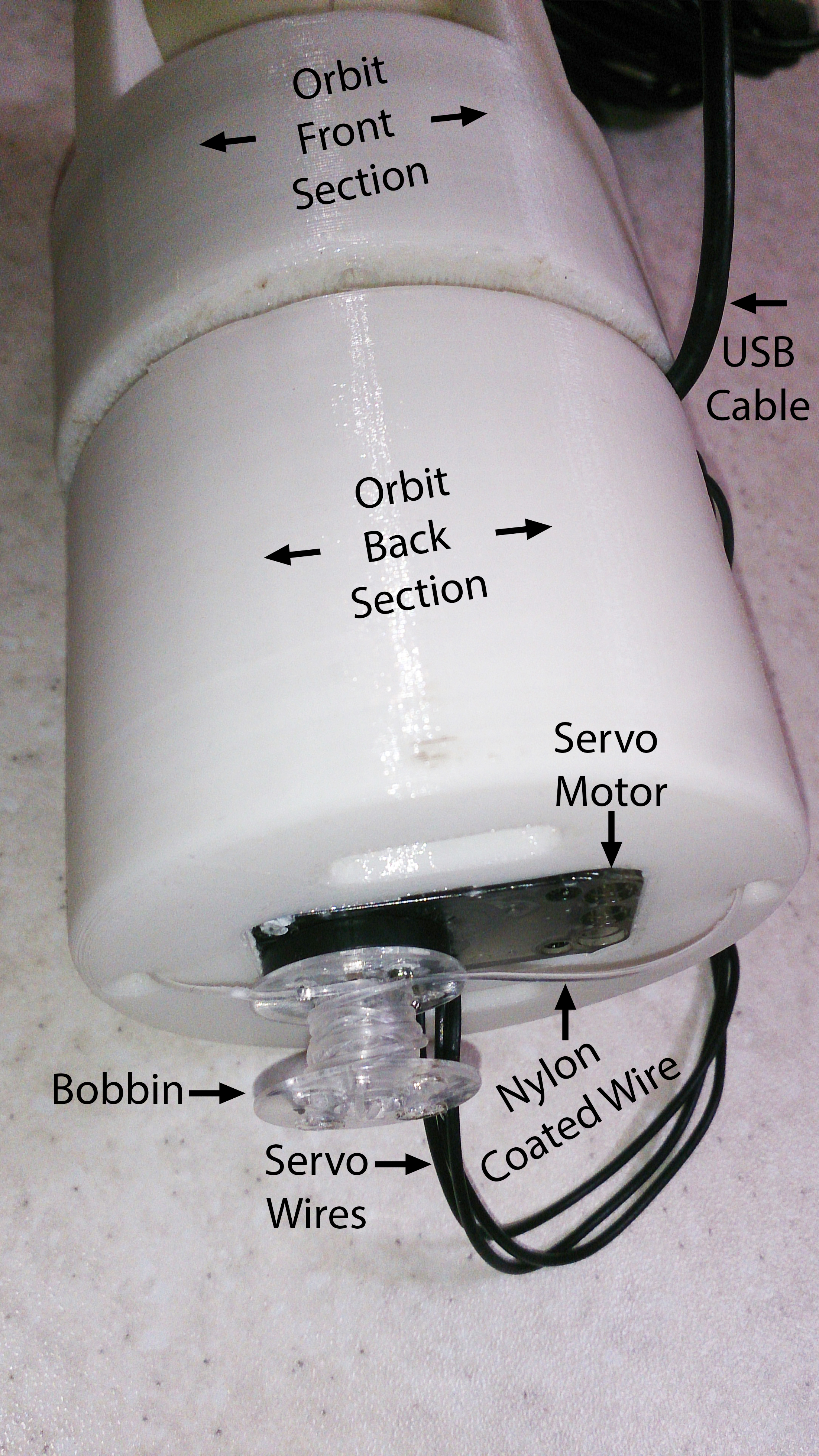}}
	\caption{The servo motor mounted within the back section of the robotic eye's orbit, as well as the woven bobbin with nylon-coated wires that are placed over the servo motor}
	\label{fig:bobbin}
\end{figure}

\begin{figure}[t]
\centerline{\includegraphics[trim=10cm 27cm 10cm 0, clip, width=\columnwidth]{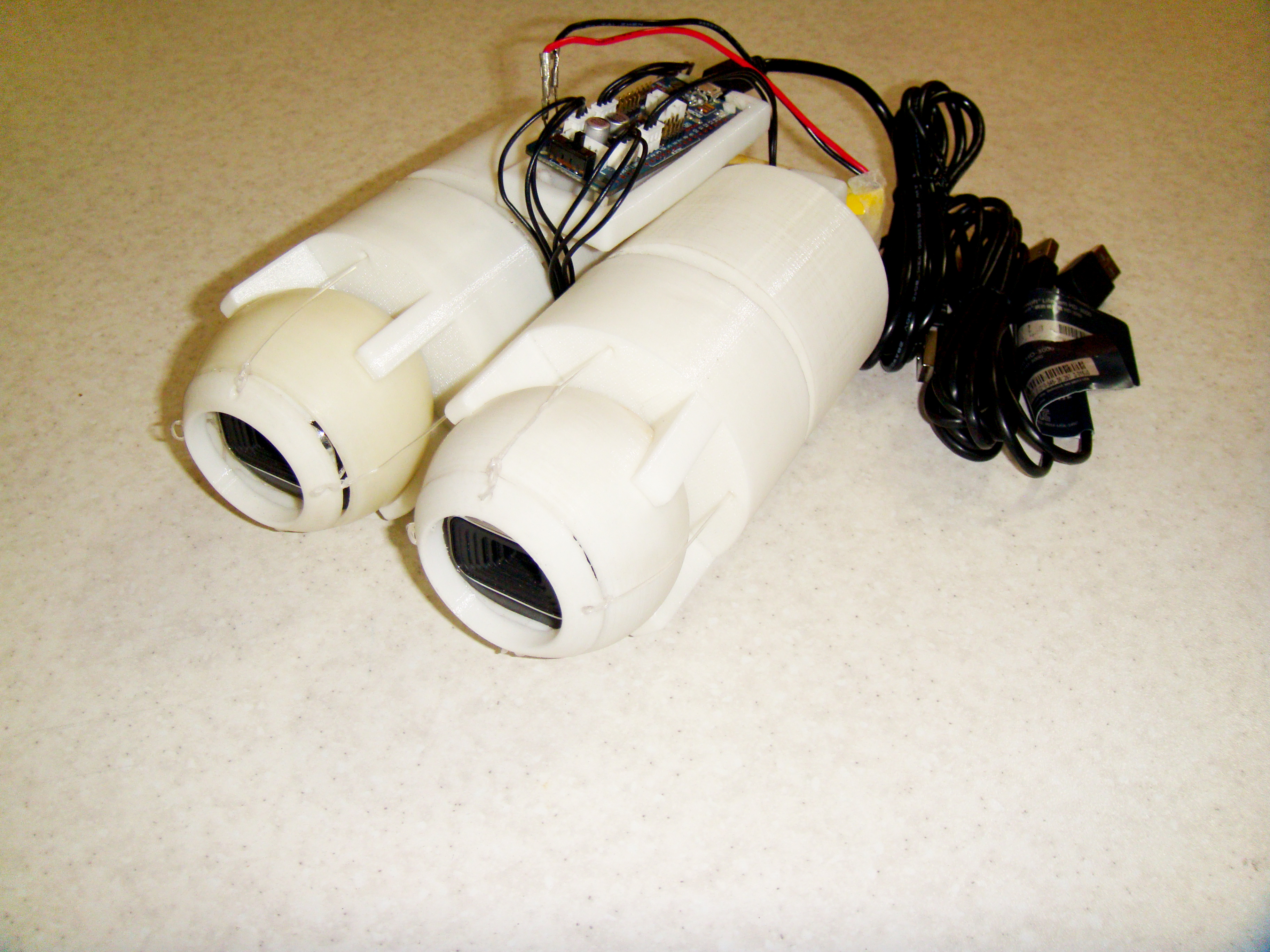}}
	\caption{The proposed robotic eye after final assembly. The final design is compact and affordable, allowing for easy replication and modification}
	\label{fig:robot}
\end{figure}

\section{Experiments and Results}

In this section, we describe the experiments conducted to evaluate the performance of our proposed robotic eye in response to changes in the environment, also known as active vision~\cite{b11wang2013advanced}. 

\begin{figure}[t]
\centering
\includegraphics[width=\columnwidth]{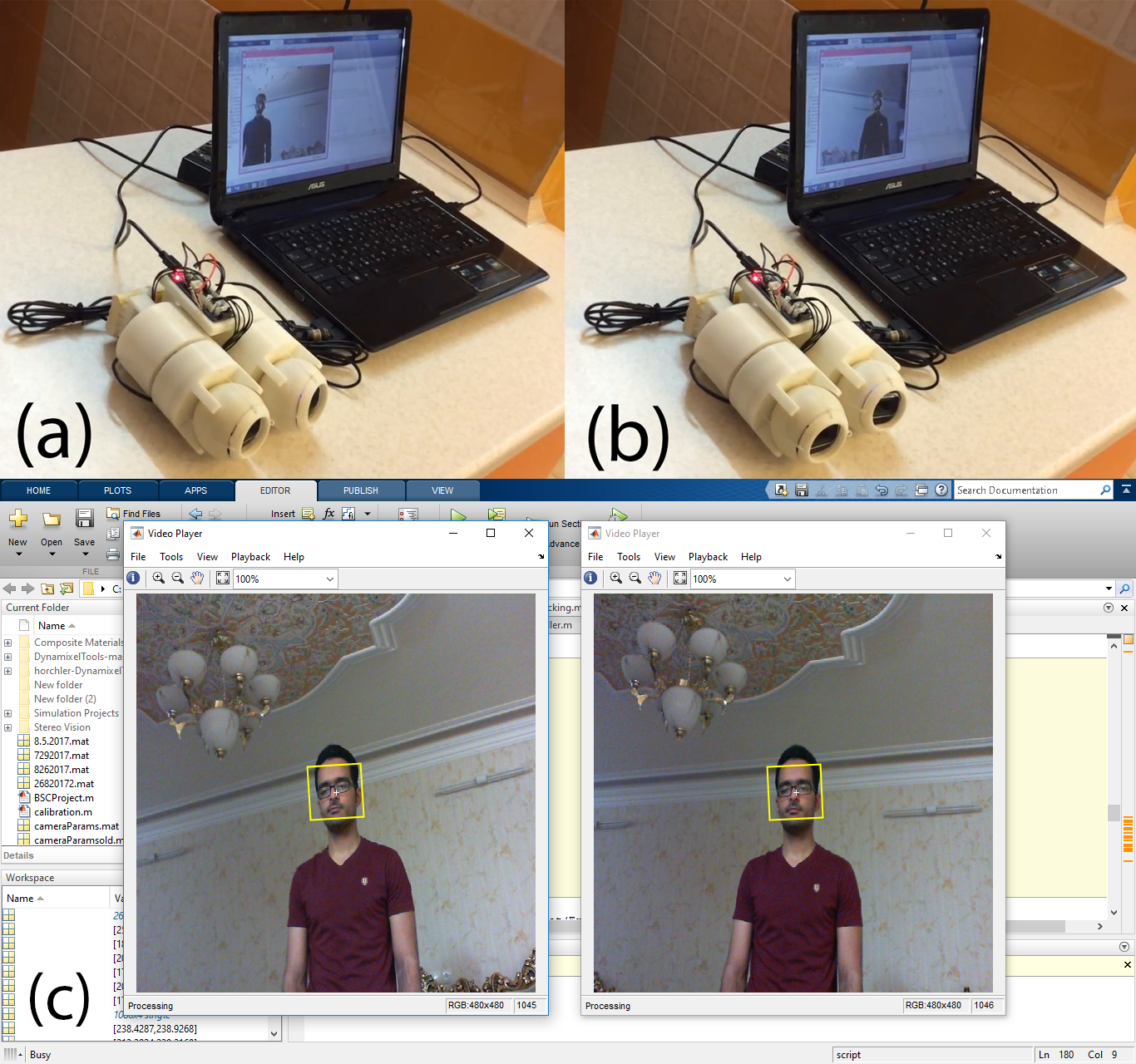}
	\caption{The proposed robotic eye during tracking the test subject's face. The figure includes three subfigures, c shows the face detection and (a)-(b) show the face following, with the face center being highlighted}
	\label{fig:face}
\end{figure}

We evaluate different eye movements independently by testing the performance of our design using a face tracker in the MATLAB environment. In all experiments, a human subject was present in front of the robot and the robot cameras captured the video stream of the scene. A face tracking algorithm in the MATLAB environment analyzes the video streams to determine the position of the test subject's face. As shown in Fig.~\ref{fig:face} The datapoints used in the controllers include the central point of the detected face bounding box.  

\begin{figure}[t]
\centering
\includegraphics[trim=0 6em 0 0, clip, width=\columnwidth]{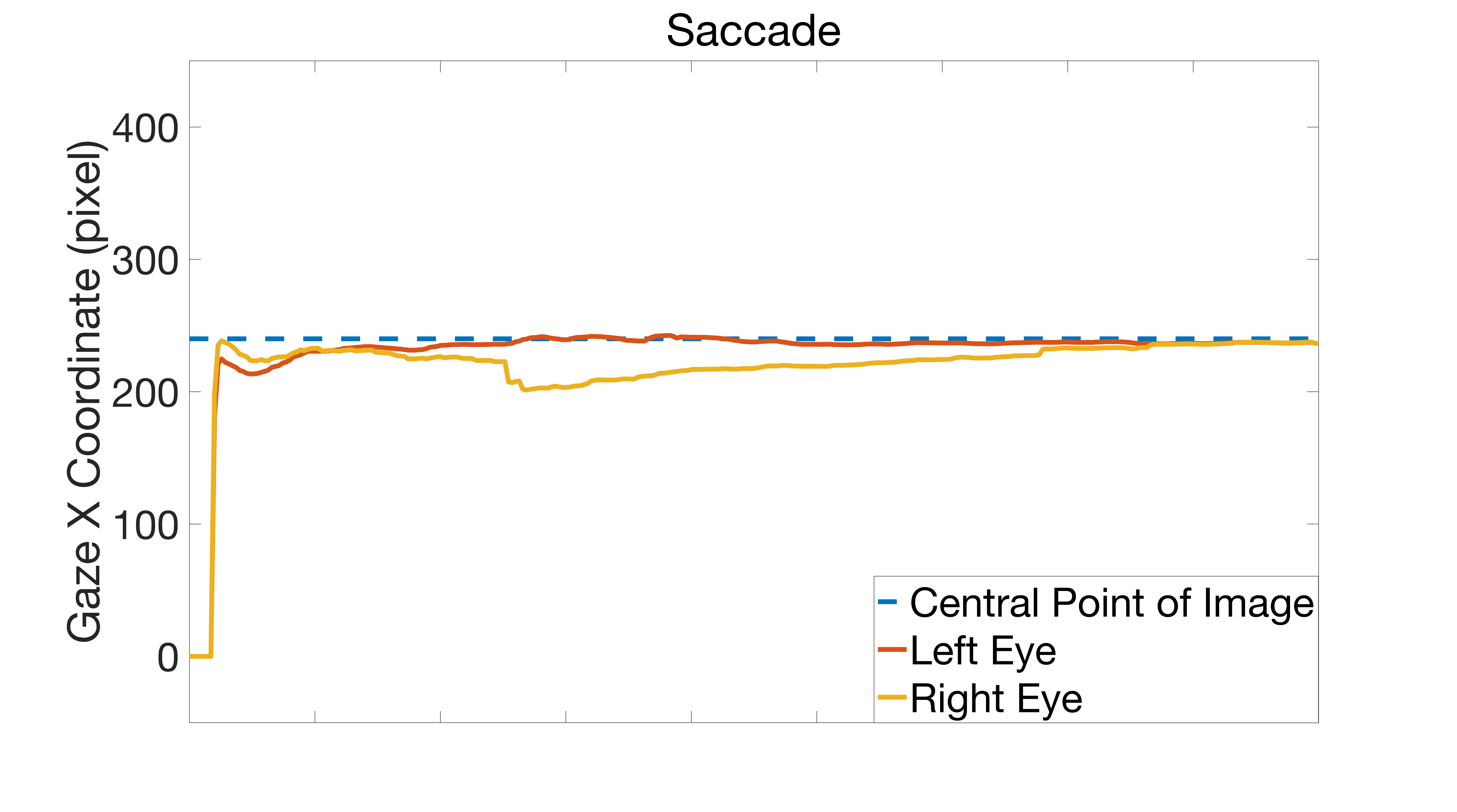}
\includegraphics[trim=0 6em 0 6em, clip, width=\columnwidth]{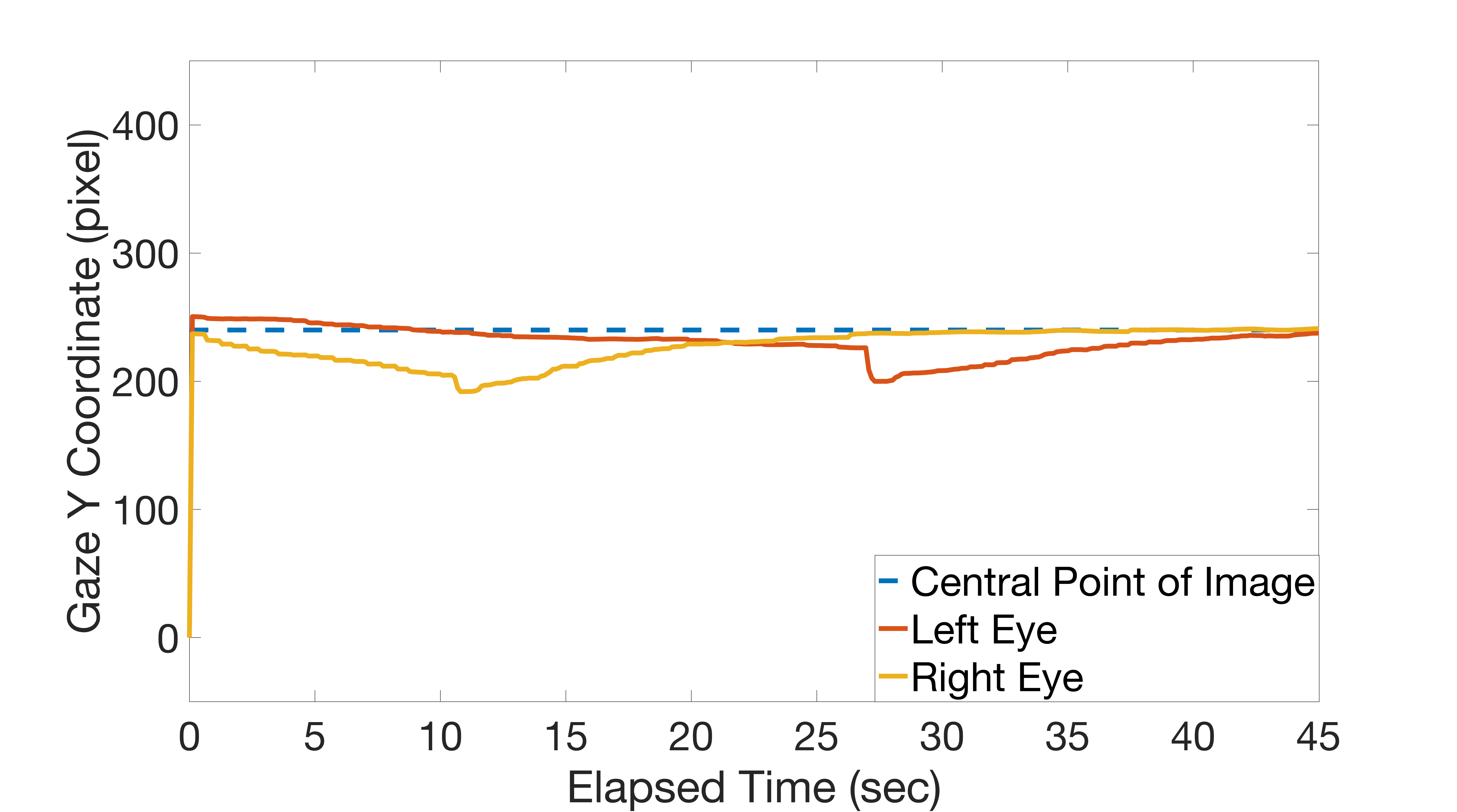}
	\caption{Saccadic evaluation results. The graphs show the X and Y coordinates of the central point of the human face as captured by the left and right cameras of the robot.}
	\label{fig14_r1}
\end{figure}

\begin{figure}[H]
\centering
\includegraphics[trim=0 6em 0 0, clip, width=\columnwidth]{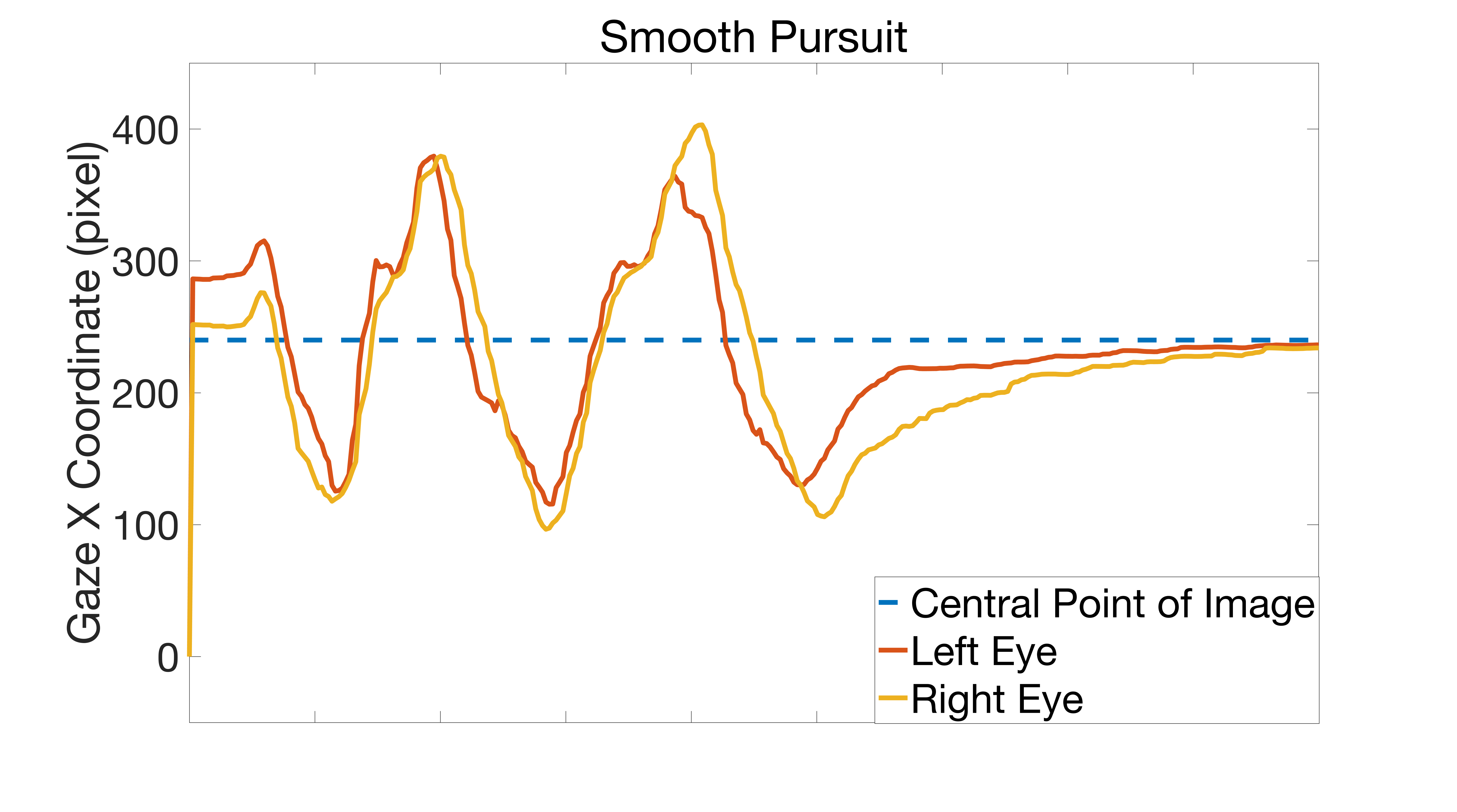}
\includegraphics[trim=0 6em 0 6em, clip, width=\columnwidth]{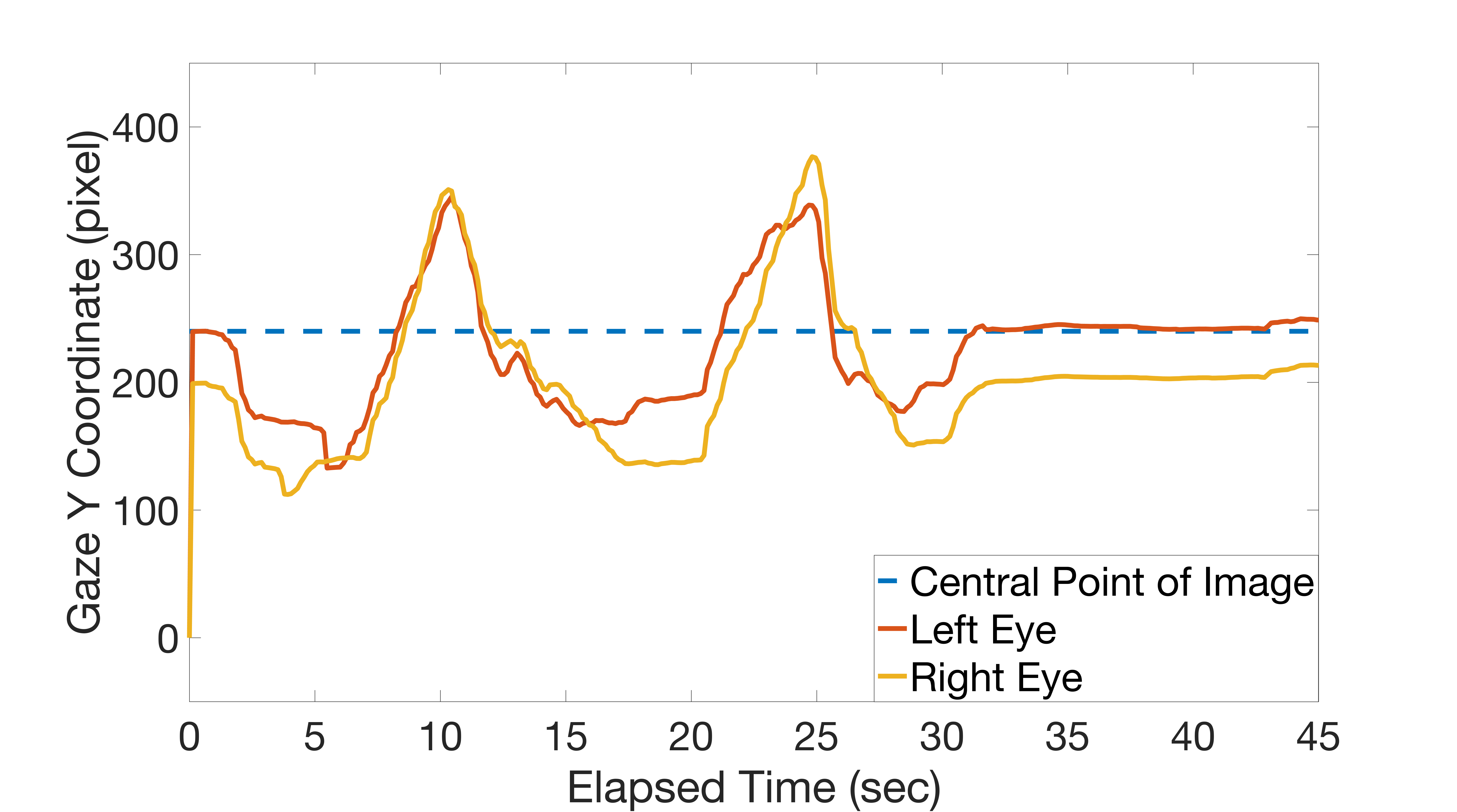}
	\caption{Smooth Pursuit evaluation results. The graphs show the X and Y coordinates of the central point of the human face as captured by the left and right cameras of the robot.}
	\label{fig14_r2}
\end{figure}

\begin{figure}[t]
\centering
\includegraphics[trim=0 6em 0 0, clip, width=\columnwidth]{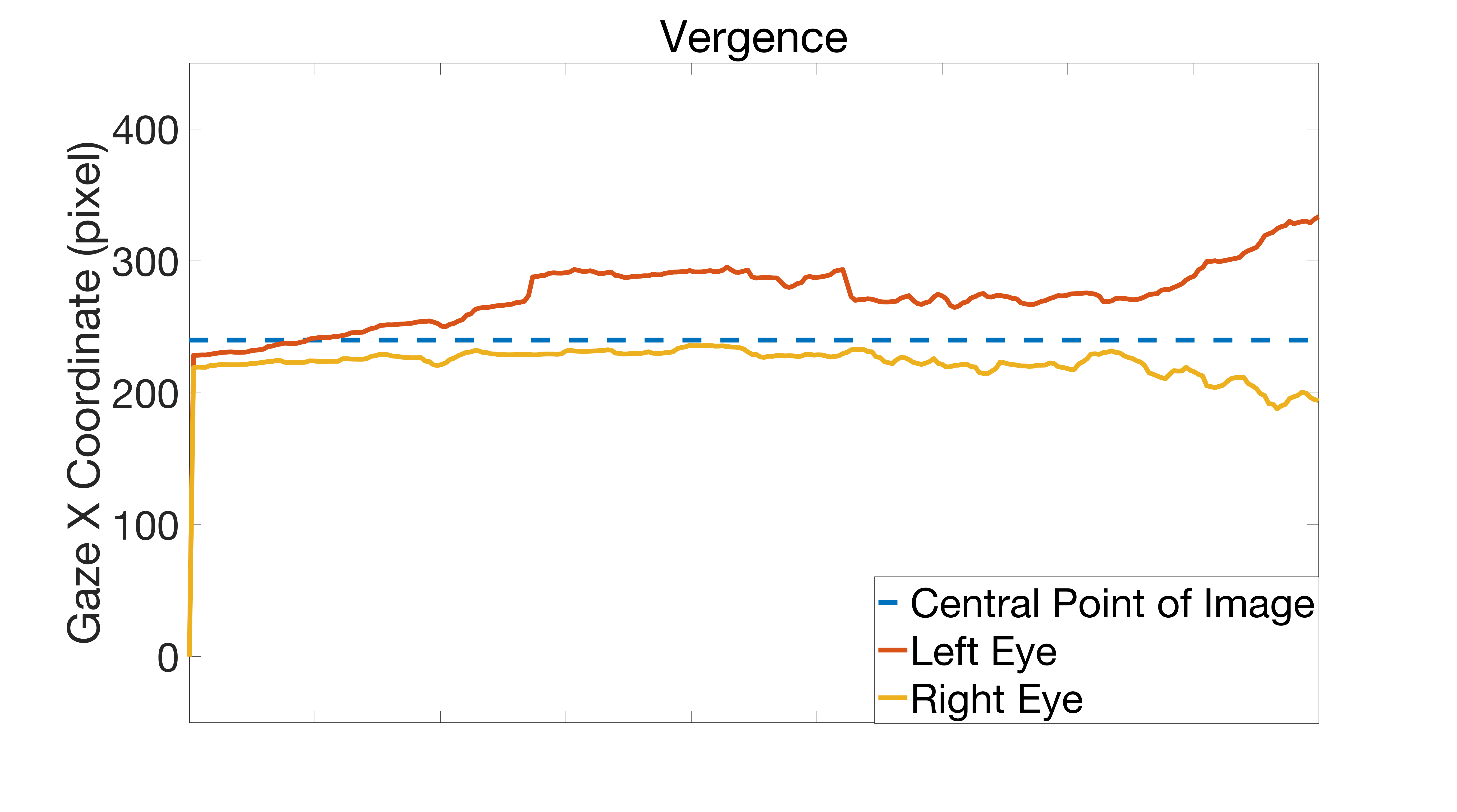}
\includegraphics[trim=0 6em 0 6em, clip, width=\columnwidth]{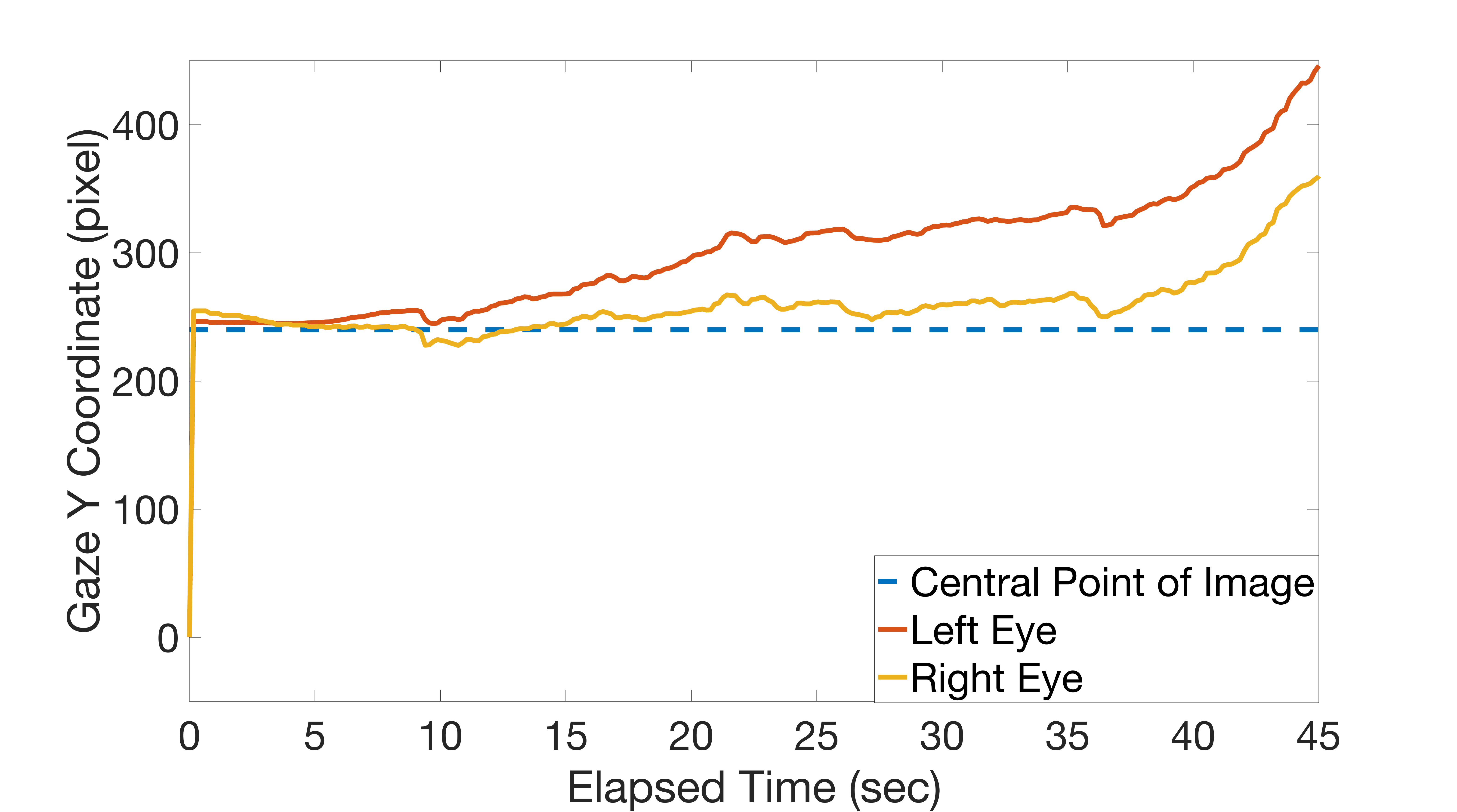}
	\caption{Vergence evaluation results. The graphs show the X and Y coordinates of the central point of the human face as captured by the left and right cameras of the robot.}
	\label{fig14_r3}
\end{figure}

\begin{figure}[t]
\centering
\includegraphics[trim=0 6em 0 0, clip, width=\columnwidth]{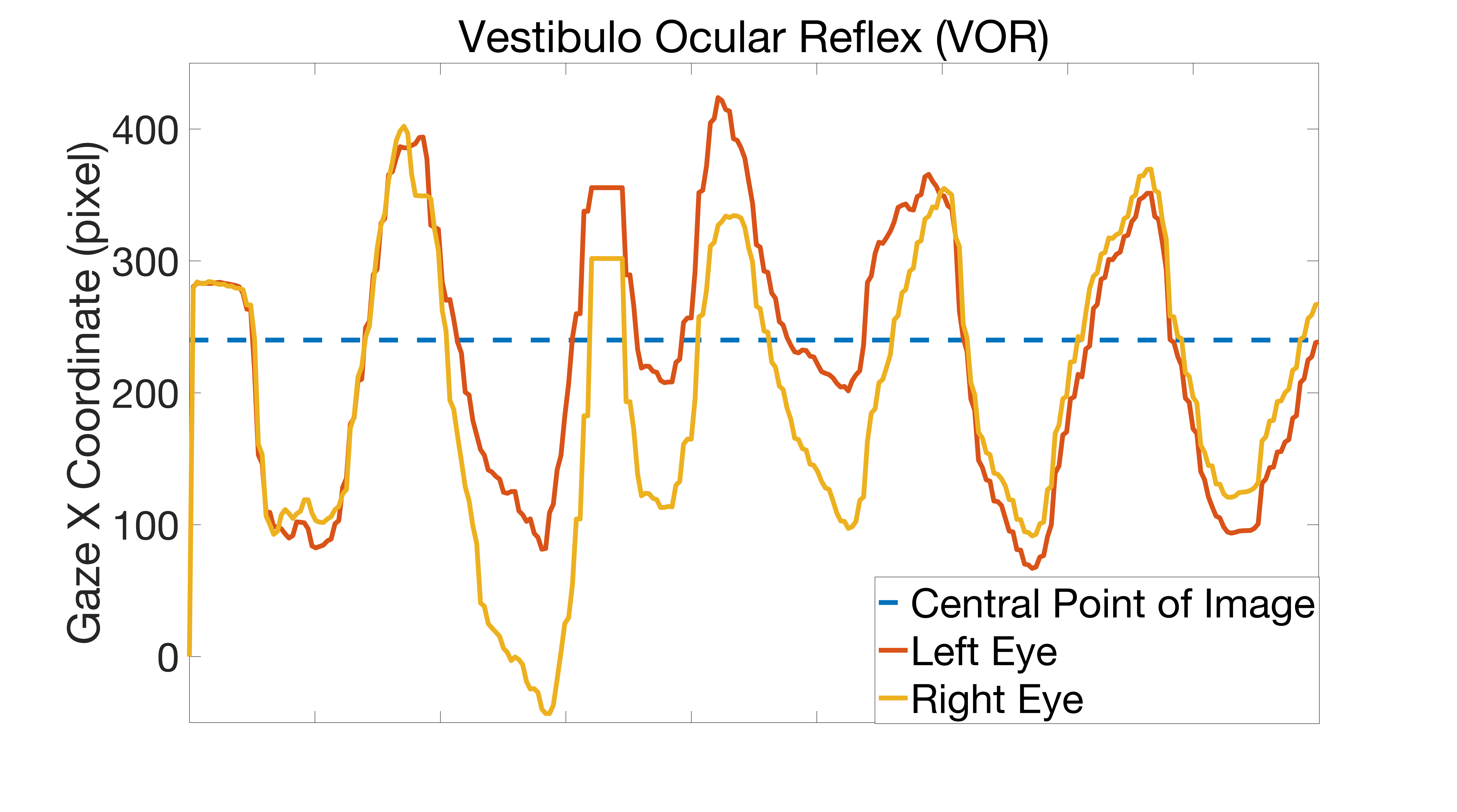}
\includegraphics[trim=0 6em 0 6em, clip, width=\columnwidth]{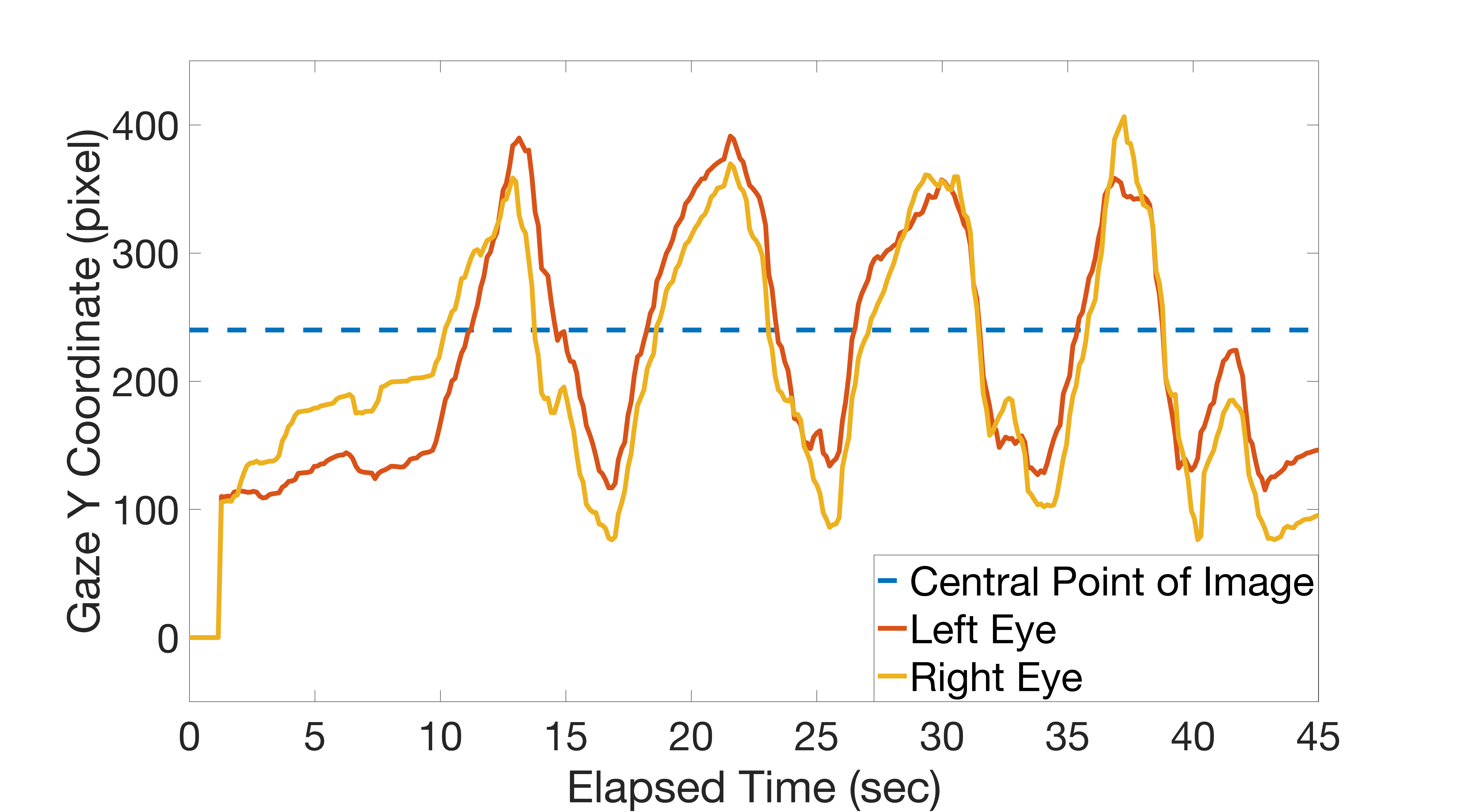}
	\caption{VOR evaluation results. The graphs show the X and Y coordinates of the central point of the human face as captured by the left and right cameras of the robot.}
	\label{fig14_r4}
\end{figure}

\subsection{Assembling the Robotic Eye}

Here we  describe the process of constructing and assembling the proposed robotic eye. The design of the eyeball and orbit allows for their use in both the left and right eyes. We then proceed to 3D print two sets of eyeball, eyeball cap, front and back components of the orbit. The order of assembly is illustrated in Fig.~\ref{fig:exploded}, and the placement of the wires and bobbins on the servos can be seen in Fig.~\ref{fig:bobbin}.

The robot uses four Dynamixel XL-320 servos as actuators to move the two cameras. After each servo is mounted in its designated chamber within the orbit, a bobbin should be placed on top and the wires should be woven around it. Once both eyes have been assembled, they are integrated into the system using the handle. 

The robot servos are connected to an OpenCM 9.04-C controller board, which receives commands from the computer through MATLAB. We made a MATLAB toolbox, XL-320, specifically for the actuators of the robotic eye to generate and simulate movements in MATLAB. The toolbox together with the code for the experiments, and the printable 3D sketches of the proposed robotic eye have been made open-source and publicly available \footnote{Code and Sketches at: \url{https://github.com/hamidosooli/robotic_eye}}.

To support all four servos simultaneously, the controller board requires an additional power source. Therefore, we added a 7.2V battery to supplement the voltage through the USB port. Fig.~\ref{fig:robot} shows the assembled robot.

To provide feedback from the environment, we used a vision-based Proportional Integral Derivative (PID) controller. The specific objective of this controller is to keep the detected face at the center of the image. We performed separate horizontal and vertical simulations for Saccade, Smooth Pursuit, Vergence, and Vestibulo-Ocular Reflex movements. The accompanying video shows our experiments\footnote{Accompanying video: \url{https://youtu.be/1mQl93-Jzi8}}.

\subsection{Saccadic Movement}
Saccadic movements are rapid eye movements that allow the eyes to quickly shift the gaze from one point to another. To evaluate our system on Saccadic movements, the human subject remained static in front of the robot at a fixed distance. The human's face was initially located near the edge of the image, simulating the scenario where the eyes have to move a long distance towards the target. 
The graphs in Fig.~\ref{fig14_r1} show that the robot's cameras move rapidly towards the central point of the human face, mimicking the Saccadic eye movement. However, upon analysis of the results, we found that the proposed robot's movements speed were not as fast as those of the human eye for this movement. This can be seen in the accompanying video. This limitation may be due to the mechanical constraints of the robot's actuators or the control algorithms used. Further research is needed to speed up the robot's movements.

\subsection{Smooth Pursuit Movement}
The Smooth Pursuit movement is used to track a moving target with smooth, continuous eye movements. To evaluate this, we conducted an experiment where a test subject walks or moves up and down while the robotic eye detects and tracks the face. Since the robot's cameras were programmed to follow the movement of the human face and maintain it at the center of the image, careful attention was given to maintaining gentle movements to ensure that the robot's movements are smooth and continuous. The results for this test were captured and plotted in Fig.~\ref{fig14_r2}.


\subsection{Vergence Movement}
Vergence movements are used to adjust the eyes' focus when looking at an object at varying distances. To evaluate this movement, the human approaches the robotic eye until his face is no longer at the center of the image. For this test, the robotic eye's cameras were adjusted to track the test subject's movement and keep their face in focus. More specifically, the robot's cameras were programmed to mimic the Vergence movement by rotating towards each other as the test subject approached. This scenario happens when the eyes have to adjust their focus to maintain a clear image of the face. As shown in Fig.~\ref{fig14_r3}, the eyes move away from the center and lose the central point of the test subject's face. 
\subsection{Vestibulo-Ocular Reflex (VOR) Movement}
Vestibulo-ocular reflex (VOR) is a reflex eye movement that compensates for head movements in one direction by moving the eyes in the opposite direction~\cite{b27barr1976voluntary}. During this experiment, the test subject was staying static while the robot was moved in different directions. The robot's cameras were programmed to mimic the VOR movement by moving in the opposite direction of the robot's body movements. This scenario happens when the head is rotated, and the VOR reflex causes the eyes to move in the opposite direction. The results of this experiment have been shown in Fig.~\ref{fig14_r4}.
\

\section{Conclusion}
We designed and evaluated a robotic eye that is capable of mimicking the main human eye movements. The robot's design was based on Computer-Aided Design (CAD) and 3D printing, which resulted in an accurate and affordable prototype. Our design uses inexpensive components, 3D printed parts, and minimum number of servos to achieve its functionality. We evaluated our robotic eye over the main four movements including Saccades, Smooth Pursuit, VOR and Vergence. As our results show, the proposed design is capable of mimicking human eye movements to a certain extent. However, there are still some limitations that need to be addressed in future work. For instance, our design does not consider movements such as torsion and intorsion. Our proposed robot can be easily modified to simulate these movements by adding an extra servo for each eye. This will allow the robot to mimic the movements of the oblique muscles in addition to those of the rectus muscles. 
The designed robot has a great potential in human-robot interaction, computer vision and cognitive science applications. To encourage further research in this area, we have made our design and code publicly available. 

\addtolength{\textheight}{-12cm}   

\bibliographystyle{IEEEtran}
\bibliography{bibtex/bib/root}

\end{document}